\documentclass[a4paper]{article}

\usepackage{INTERSPEECH2021}
\usepackage{multirow}
\usepackage{color}
\usepackage[hyphens]{url}
\usepackage[hidelinks]{hyperref}
\usepackage[hyphenbreaks]{breakurl}
\usepackage[capitalize]{cleveref}
\usepackage{enumitem}
\usepackage{etoolbox}
\usepackage{graphics}

\newcommand{\asrProb}{\ensuremath{P_{\mathrm{AED}}}}
\newcommand{\lmProb}{\ensuremath{{P_{\mathrm{LM}}}}}

\newcommand{\ilmProbExact}{\ensuremath{{P_{\mathrm{ILM}}}}}
\newcommand{\ilmProbGeneric}{\ensuremath{{\hat{P}_{\mathrm{ILM}}}}}
\newcommand{\ilmProbExt}{\ensuremath{{P_{\mathrm{train-LM}}}}}

\newcommand{\ilmIdxZero}{\ensuremath{\operatorname{zero}}}
\newcommand{\ilmIdxCtxAvg}{\ensuremath{\EX_{\mathcal{D}}[c]}}
\newcommand{\ilmIdxEncAvg}{\ensuremath{\EX_{\mathcal{D}}[h]}}
\newcommand{\ilmIdxSeqEncAvg}{\ensuremath{\EX_{x}[h]}}

\newcommand{\ilmIdxTrainedLstm}{\ensuremath{\operatorname{Mini-LSTM}}}

\newcommand{\ilmProbZero}{\ensuremath{{\hat{P}_{\mathrm{ILM}_{\ilmIdxZero}}}}}

\newcommand{\encDim}{\ensuremath{D_{\textrm{enc}}}}

\DeclareMathOperator*{\argmax}{arg\,max}

\DeclareMathOperator{\EX}{\mathbb{E}}

\newcommand{\R}{\mathbb{R}}

\renewrobustcmd{\bfseries}{\fontseries{b}\selectfont}
\renewrobustcmd{\boldmath}{}
\newrobustcmd{\B}{\bfseries}

\newsavebox\CBox

\makeatletter

\renewcommand{\section}{\@startsection
   {section}%
   {1}%
   {}%
   {-0.4\baselineskip}%
   {0.2\baselineskip}%
   {}}%

\renewcommand{\subsection}{\@startsection
  {subsection}%
  {2}%
  {}%
  {-0.1\baselineskip}%
  {0.1\baselineskip}%
  {}}%

\renewcommand{\subsubsection}{\@startsection
  {subsubsection}%
  {3}%
  {}%
  {-0.2\baselineskip}%
  {0.2\baselineskip}%
  {}}%

\g@addto@macro\normalsize{%
  \setlength\abovedisplayskip{3pt plus 2pt minus 1pt}
  \setlength\belowdisplayskip{3pt plus 2pt minus 1pt}
  \setlength\abovedisplayshortskip{2pt plus 2pt minus 1pt}
  \setlength\belowdisplayshortskip{2pt plus 2pt minus 1pt}
}

\setlist{
        itemsep=0pt,
        parsep=1pt plus 1pt minus 1pt,
        topsep=1pt plus 1pt minus 1pt,
        partopsep=0pt}

\title{Investigating Methods to Improve Language Model Integration for
  Attention-based Encoder-Decoder ASR Models}
\name{Mohammad Zeineldeen$^{1,2}$, Aleksandr Glushko$^1$, Wilfried Michel$^{1,2}$, Albert Zeyer$^{1,2}$, Ralf Schlüter$^{1,2}$, Hermann Ney$^{1,2}$}
\address{
  $^1$Human Language Technology and Pattern Recognition,
  Computer Science Department, RWTH Aachen University, 52074 Aachen, Germany \\
  $^2$AppTek GmbH, 52062 Aachen, Germany}
\email{\footnotesize \{zeineldeen, michel, zeyer, schlueter, ney\}@cs.rwth-aachen.de, aleksandr.glushko@rwth-aachen.de}

\begin{document}

\maketitle
\begin{abstract}
Attention-based encoder-decoder (AED) models learn an implicit internal
language model (ILM) from the training transcriptions.
The integration with an external LM trained on much more unpaired text
usually leads to better performance.
A Bayesian interpretation
as in the hybrid autoregressive transducer (HAT)
suggests dividing by the prior of
the discriminative acoustic model,
which corresponds to this implicit LM,
similarly as in the hybrid hidden Markov model approach.
The implicit LM cannot be calculated efficiently in general and
it is yet unclear what are the best methods to estimate it.
In this work, we compare different approaches from the literature
and propose several novel methods to estimate the ILM directly from the AED
model.
Our proposed methods outperform all previous approaches.
We also investigate other methods to suppress the ILM mainly by decreasing
the capacity of the AED model, limiting the label context, and
also by training the AED model together with a pre-existing LM.
\end{abstract}

\section{Introduction \& Related Work}

End-to-end automatic speech recognition (ASR) models have shown competitive
results on a variety of tasks \cite{park2019specaugment, Li2020DevelopingRM}.
One of the most popular models is the attention-based encoder-decoder (AED)
model \cite{chan2016las}.
This model can learn a direct mapping from input features to output labels
by jointly learning the acoustic model (AM), pronunciation model,
and language model (LM).
The encoder maps the input features into some high-level representations.
The decoder acts as a language model which uses attention to summarize the
encoder's representations to produce output labels.

Monolingual or unpaired text data is usually much larger in magnitude than
transcribed paired data.
Therefore, using an external LM trained on monolingual text data
in combination with the ASR model often yields
better performance.
There have been many studies on how to integrate the ASR model
with an external LM.
Shallow fusion \cite{lmMethodsCompare} is often used which is simply a
log-linear interpolation of the ASR and LM scores during inference.
Other, more sophisticated approaches including Cold Fusion
\cite{Sriram2018ColdFT}, Deep Fusion \cite{Glehre2015DeepFT},
Simple Fusion \cite{stahlberg2018SimpleFT}, and
Component Fusion \cite{shan2019CompFT} fuse the LM into the ASR
model by combining their hidden states during training.
There also have been studies to include the LM into the training criterion in a
log-linear fashion \cite{michel2020localfusion}.
Among these studies, shallow fusion seems to be the most dominantly used approach.

Because of the discriminative ASR model,
a Bayes interpretation suggests dividing by the prior of the ASR model
when combining with an external LM.
During inference, the log-domain score of this prior model is subtracted
from the log-linear combination of both ASR model and external LM scores.
Due to the context-dependency on previous labels,
the ASR model has implicitly learned an \emph{internal LM} (ILM),
which corresponds to this prior model.
The internal LM can not be calculated exactly
but there are various ways how to approximate it.
In case of AED or RNN-T \cite{Graves2012SequenceTW},
the internal LM is assumed to be mostly learned by the decoder or prediction network,
and various simple ways exist for its estimation
by masking out the encoder.
This was first suggested as part of the hybrid autoregressive transducer (HAT)
\cite{variani2020hat}.
The Density Ratio (DR) approach \cite{McDermott2019DensityR}
is similar.
It does not estimate the internal LM directly
but instead estimates the prior as a separate LM
trained on the training transcriptions only.
Subtracting the internal LM during inference can be seen as removing the prior
or bias of the training text data.

The authors of \cite{Meng2020InternalLM} also proposed a simple way to estimate the internal LM
for AED models inspired by HAT, which is masking out the attention context vector.
We call this \text{\ilmIdxZero} approach.
We argue that this was applied under a strong assumption that the AED decoder
modeling also follows the proof proposed by HAT (\cite{variani2020hat} Appendix A).
Instead, we believe that other ILM estimation methods can be more accurate when
considering the encoder bias information as input instead of just zero.

Furthermore, it was observed that the effect of an external LM can be smaller
if the decoder is over-parameterized \cite{Tske2020SingleHA}.
This motivated us to study different ways of suppressing the ILM of the decoder.
For AED models, this can be done by weakening the decoder by decreasing its
number of parameters or accessible label context or by including an LM
in the training process.

In this work, we present several novel methods to estimate the ILM directly
from the AED model.
We propose methods that use the averaged encoder states or
averaged attention context vectors over all training data instead of zero input.
We also introduce a new method called \ilmIdxTrainedLstm, which is trained
to minimize directly the perplexity of the ILM and achieved the best results.
We show that all our methods outperform the previously proposed methods namely
Shallow Fusion, Density Ratio, and ILM estimation with \text{\ilmIdxZero} encoder \cite{Meng2020InternalLM}.
Moreover, we show that it is possible to train an AED model with a limited
context decoder and still achieve comparable results with an LSTM decoder
on some tasks.

\section{Model}

Our model follows the attention-based encoder-decoder model
\cite{zeyer2018:asr-attention,zeyer2019:trafo-vs-lstm-asr}.
The encoder maps the input sequence $x_1^{T'}$ into a sequence of
hidden states $h_1^T$ where $T' \ge T$ due to down-sampling,
and $h_t \in \R^{\encDim}$.
For each decoder step $i$, the attention mechanism is used to calculate
attention weights $\alpha_{i,t}$ as:
\begin{equation}
\label{att:weigths}
    \alpha_{i,t} := \mathrm{Softmax}(\mathrm{Attention}(s_i, \beta_{i,t}, h_1^T))
\end{equation}
where $\beta_{i,t}$ is attention location-awareness feedback.
The attention context vector $c_i \in \R^{\encDim}$ is then computed as a weighted sum over the encoder hidden states $h_1^T$ as
\begin{equation}
\label{att:context}
    c_i := \sum_{t = 1}^T \alpha_{i, t} h_t, \quad\quad c_0 := 0 .
\end{equation}
The decoder state is modeled as a recurrent function using LSTM:
\begin{equation}
\label{att:query}
    s_i := \operatorname{LSTM}(s_{i - 1}, y_{i - 1}, c_{i - 1})
\end{equation}
Finally, the output probability for some label $y_i$ is computed as:
\begin{equation}
    p(y_i | y_{i - 1}, x_1^T) = \operatorname{Softmax}(\mathrm{MLP}_{\mathrm{readout}}(s_i, y_{i - 1}, c_i))
\end{equation}
where $\operatorname{MLP}_{\mathrm{readout}} = \operatorname{linear} \circ \operatorname{maxout}
\circ \operatorname{linear}$.

\subsection{Feed-forward (FF) Decoder}

The decoder state is usually recurrent and modeled by an LSTM allowing full
label context feedback.
However, we can also limit the label context by modeling the decoder state
using feed-forward (FF) layers.
The computation of the decoder state is then
defined as:
\begin{equation}
\label{eqn:FFDecoder}
    s_i = \mathrm{FF}(y_{i - k}^{i - 1}, c_{i - 1})
\end{equation}
where $k$ is the context size.
The decoder state $s_i$ has no explicit dependency on the
previous state $s_{i-1}$ but an implicit one via the attention
mechanism in \Cref{att:weigths} and \Cref{att:context}.

\section{Internal LM Estimation}
\label{sec:ILM}

During inference, the search algorithm searches for the best word sequence $w_1^N$ that maximizes:
\begin{equation*}
\hat{w}_1^{\hat{N}} = \argmax_{N, w_1^N} \left\lbrace \log P(w_1^N | x_1^T ) \right\rbrace
\end{equation*}
In principle, the sentence posterior probability can be modeled directly by
the AED model.
In practice, however, significant improvement can be
obtained by including an external LM via log-linear model
combination (\textit{shallow fusion}) \cite{lmMethodsCompare}.
During training, the AED model learns an internal LM (ILM) from the training data.
It has been shown that additional benefit can be obtained when the ILM
sequence prior information can be removed \cite{variani2020hat,Meng2020InternalLM}.

Hence our probability is modeled by a combination of the three models
\begin{equation*}
P(w_1^N \vert x_1^T ) \propto \asrProb (w_1^N \vert x_1^T) \cdot  P_{\mathrm{LM}}^{\lambda_1} (w_1^N) \cdot P_{\mathrm{ILM}}^{- \lambda_2} (w_1^N)
\end{equation*}
where $\asrProb$ is the AED model probability, $\lmProb$ and $\ilmProbExact$ are
the external LM and ILM respectively and $\lambda_1$ and $\lambda_2$
are scalar model scales.

The estimation of the ILM is not straightforward as it is implicitly modeled
in the decoder.
The ILM is defined as:
\begin{equation}
    \ilmProbExact (w_1^N) = \sum_{T,x_1^T} \asrProb (w_1^N \vert x_1^T) \cdot P(x_1^T)
    \label{eqn:exactILM}
\end{equation}
However, the summation is intractable.
%
In the following, we will propose different estimation methods by modifying
the attention context vector $c_i$ since it represents the input to the
decoder in our case,
as an approximation to estimate $\ilmProbExact$.
\begin{equation}
\ilmProbExact (w_1^N) \approx \ilmProbGeneric (w_1^N) := \asrProb (w_1^N \vert c_i = \hat{c}_i)
\end{equation}

\subsection{Zero-Attention ILM Estimation}
We set the context vector to zero as
proposed by \cite{variani2020hat,Meng2020InternalLM}.
$\ilmProbZero$ is defined by
\begin{equation*}
  \hat{c}^{\ilmIdxZero}_i := 0 \quad \forall i .
\end{equation*}

\subsection{Average-Attention ILM Estimation}

The \text{\ilmIdxZero} ILM estimation
removes all encoder bias from the decoder.
However, if this bias appears consistently in the encoder information, it can be
considered part of the ILM and should also be accounted for during inference.
To also capture this bias, we use the average encoder state instead of zero.
There are several ways to estimate the average

\textbf{Global Attention Context Average (\ilmIdxCtxAvg)}.
In this approach we average the attention context vectors $c$
over all examples of
our training data $(x,y) \in \mathcal{D}$ with $c_j(x,y)$:
\begin{equation*}
  \hat{c}_i^{\ilmIdxCtxAvg} :=
  \frac{1}{J_{\mathrm{tot}}}
  \sum_{(x,y_1^J) \in \mathcal{D}}
  \sum_{j=1}^J c_j (x,y) \quad \forall i \ge 1,
  \quad \hat{c}_0^{\ilmIdxCtxAvg} := 0,
\end{equation*}
and $J_{\mathrm{tot}} := \sum_{(x,y_1^J) \in \mathcal{D}} J$.

Note that we also tested not using the special case for $i=0$.
This performed slightly worse in some cases,
as it becomes slightly more inconsistent
because $c_0 := 0$ is exactly what we defined for the AED model
(\Cref{att:context}).

\textbf{Global Encoder Average (\ilmIdxEncAvg)}.
Instead of taking the average context vector we can also directly average the
encoder output $h$ over the entire training data.
\begin{equation*}
  \hat{c}_i^{\ilmIdxEncAvg} :=
  \frac{1}{T_{\mathrm{tot}}}
  \sum_{(x_1^{T_x},y) \in \mathcal{D}}
  \sum_{t=1}^{T_x} h_t (x) \quad \forall i \ge 1,
  \quad \hat{c}_0^{\ilmIdxEncAvg} := 0
\end{equation*}
and $T_{\mathrm{tot}} := \sum_{(x_1^{T_x},y) \in \mathcal{D}} T_x$.

\textbf{Sequence-Level Encoder Average (\ilmIdxSeqEncAvg)}.
To capture gradual shifts in the encoder imposed bias we restricted
the average to the current utterance $(x,y)$.
\begin{equation*}
  \hat{c}_i^{\ilmIdxSeqEncAvg} := \frac{1}{T} \sum_{t=1}^{T} h_t(x) \quad \forall i \ge 1,
  \quad\quad \hat{c}_0^{\ilmIdxSeqEncAvg} := 0
\end{equation*}
Note that this is not a proper ILM estimation as it makes uses of the input $x$.

\subsection{Trained Context ILM Estimation}

In this approach,
we define $\hat{c}$ by a mini model which is trained to minimize the perplexity
of the ILM, while keeping all other ASR model parameters fixed.


\textbf{Mini-LSTM}.
We can use a mini LSTM defined as
\begin{equation*}
\hat{c}_i^{\ilmIdxTrainedLstm} :=\operatorname{linear} \circ \operatorname{LSTM}(y_1^{i-1},\theta_{\ilmIdxTrainedLstm}) \quad \forall i
\end{equation*}
with trainable parameters $\theta_{\ilmIdxTrainedLstm}$.
Specifically, we use the same decoder embedding for $y$ as input to the LSTM
consisting of 50 hidden units and then projected to $\R^{\encDim}$.
The training steps are summarized as follows:
(1) freeze trained $\theta_{\text{AED}}$,
(2) replace $c_i$ with $\hat{c}_i^{\ilmIdxTrainedLstm}$,
(3) retrain the AED model by only updating the $\operatorname{linear}$
and Mini-LSTM parameters.

This uses only a small amount of parameters
to avoid any potential overfitting
and to avoid learning another internal LM on its own.
We train the parameters only on a subset of the training data.
This estimation is just as efficient as calculating
the average statistics above.
Loading the dataset from disk is our bottleneck.

\subsection{Decoder-Like LM}
All the above methods try to estimate an LM from the AED model
contrary to the fact that the AED model was never explicitly trained to be an LM.
A more direct approach to model the LM probability that can \textit{in principle}
be learned by the AED is to train a proxy model.

We propose to train a dedicated LM that has the same topology as the AED decoder
and is trained on the training transcriptions.
This is closely related to the density ratio approach \cite{McDermott2019DensityR}
where only source and target domain are matched in our case,
but the capacity (\# of params) of the source and target LM are differing.
\begin{equation*}
\ilmProbExact(w_1^N) \approx \ilmProbExt (w_1^N) := P_{\mathrm{decoder-like}}(w_1^N)
\end{equation*}

This method only models the theoretical LM capacity of the decoder and does
not take any specific AED model into account.

\subsection{ILM Suppression}
Instead of estimating the ILM of an existing AED model we can also try to
proactively suppress the formation of an ILM during the training such that
no or fewer corrections are necessary at decoding time.
In this work, we propose the following measures to suppress the ILM.

\textbf{Under-Parameterized Decoder}
To limit the capacity of the AED decoder we reduce the number of hidden
units in the decoder layer.
The intuition is that with reduced capacity the model has to put more
emphasis on the acoustic part to remain a viable AM and will reduce
the effort put into language modeling.
This is compensated in decoding by the strong external LM.

\textbf{Limited Context Decoder}
The power of LSTM and Transformer LMs comes from the fact that these models
have unlimited history and can effectively model the long semantic contexts
of natural languages.

We propose to replace the recurrent decoder (Equation \ref{att:query}) with
a limited context FF model (\Cref{eqn:FFDecoder}).
We argue that limiting the visible context will not reduce the acoustic
modeling capacity of the decoder while it can strongly reduce the effectiveness
as a language model.

\textbf{Train AED together with LM (train w. LM)}
Usually, the AED model and the LM are trained independently.
There are approaches to include an external LM in the training of AMs via
sequence training or local log-linear combination
\cite{michel2020localfusion}.
We argue that when an LM is present during training the AM learns to rely on the
LM for language modeling and focuses more on acoustic modeling.
Thereby, the ILM should be much weaker.

\section{Experiments}

We use RETURNN \cite{zeyer2018:returnn} for training and inference.
All LM scales $\lambda_1$ and $\lambda_2$ are tuned using grid search.
The LMs trained for Density Ratio have the same topology as the AED
LSTM-based decoder.
All our config files and code to reproduce these experiments can be found
online\footnote{\scriptsize\url{https://github.com/rwth-i6/returnn-experiments/tree/master/2021-internal-lm-attention}}.

\subsection{Switchboard 300h}

We conduct experiments on the Switchboard 300 hours dataset which consists of
English telephone conversations \cite{godfrey1992switchboard}.
We use Hub5'00 as development set which consists of Switchboard and CallHome.
We use Hub5'01 and RT03 as test sets.

We use 40-dimensional Gammatone features \cite{Schlter2007GammatoneFA}.
Our encoder consists of 2 convolutional layers followed by 6 bidirectional
LSTM layers with 1024 dimensions in each direction.
The decoder consists of a Zoneout LSTM \cite{Krueger2017ZoneoutRR} layer with
1000 dimensions.
We apply dropout of 30\% and weight decay \cite{Krogh1991ASW} of value 1e-4 for
further regularization.
We also use SpecAugment \cite{park2019specaugment} for data augmentation.
We use byte-pair-encoding (BPE) \cite{sennrich2015neuralbpe} as output labels
with a vocabulary size of 534.
All models are trained for 33 epochs.
The LM scales are tuned on Hub5'00 set and we use a beam size of 12 for inference.
The external LM is a 6-layer Transformer model based on \cite{Irie2019LanguageMW}
and trained on both transcription and Fisher data.

\subsection{LibriSpeech 960h}

We also conduct experiments on the LibriSpeech 960h
corpus \cite{Panayotov2015librispeech}.
The model architecture is similar to the one used for Switchboard, except that
we use a standard LSTM in the decoder.
We use 40-dimensional MFCC as input features and BPE as output labels with a
vocabulary size of 10k.
All models are trained for 15 epochs.
The LM scales are tuned on dev-other set and we use a beam size of 32 for
inference.
We use a 24-layer Transformer model following \cite{Irie2019LanguageMW}
as an external LM.

\subsection{ILM Estimation Methods}

We evaluate our AED model with different LM integration methods on both
Switchboard and LibriSpeech datasets.
The results are shown in \Cref{tab:swb-lstm-ilme,tab:libri-lstm-ilme}.
We observe that our proposed ILM estimation methods consistently
outperform all of the previous approaches namely
Shallow Fusion, Density Ratio, and ILM estimation with \text{\ilmIdxZero} encoder.
On Switchboard, $\ilmIdxTrainedLstm$ and $\ilmIdxCtxAvg$ methods perform the
best on Hub5'00.
With $\ilmIdxTrainedLstm$ method, we achieve $3.7\%$, $6.15\%$, and $5.7\%$
relative improvement in WER on Hub5'00, Hub5'01, and RT03 sets respectively
compared to Shallow Fusion.
On LibriSpeech, we observe even more improvement using
the $\ilmIdxTrainedLstm$ method yielding $15.3\%$ and $14.0\%$ relative
reduction in WER on dev-other and test-other respectively.
Moreover, we compute the ILM perplexities (PPLs) on the dev sets and we notice
that better methods have better PPLs.
The dev perplexity for \ilmIdxSeqEncAvg is low which might be because we
use the dev set input information during inference for the ILM.
We can also see that \text{\ilmIdxZero} is not the best estimation method which
means that using encoder information bias is helpful.

\textbf{Cross-domain inference} We also conduct experiments on a
cross-domain target dataset.
We use the AED model trained on LibriSpeech as a source-domain model and
chose TED-LIUM-V2 \cite{rousseau-etal-2014-enhancing} to be the
target-domain dataset.
TED-LIUM-V2 consists of Ted talks speech.
We trained an LSTM LM with 255M words of external target-domain text data.
We use a beam size of 12.
Results are shown in \Cref{tab:cross-lstm}.
We can observe again that $\ilmIdxTrainedLstm$ method performs the best.
It achieves $13.0\%$ and $12.4\%$ relative improvement in WER on both
dev and test sets respectively compared to Shallow Fusion.

\begin{table}[h]
    \caption{WERs [\%] on Switchboard 300h of AED model with Shallow Fusion (SF),
    Density Ratio (DR), and the proposed ILM estimation methods.
    ILM BPE-level perplexity (PPL) is reported on Hub5'00.}
    \label{tab:swb-lstm-ilme}
    \centering
    \resizebox{\columnwidth}{!}{
    \begin{tabular}{|l|c|c|c|c|c|c|}
    \hline
    Method & \shortstack{$\lambda_1$ \\ (LM)} & \shortstack{$\lambda_2$ \\ (ILM)} &
    Hub5'00 & Hub5'01 & RT03 & \shortstack{ILM \\ PPL} \\ \hline
    None &  0.0\phantom{1}  & 0.0\phantom{1} &  13.8 & 13.4 & 16.3 & - \\ \hline \hline
    SF   &  0.08 & 0.0\phantom{1} & 13.5 & 13.0 & 15.7 & - \\ \hline
    DR & 0.14 & 0.12 & 13.2 & 12.7 & 15.3 & - \\ \hline \hline
    \ilmIdxZero & 0.10 & 0.02 & 13.5 & 12.9 & 15.6 & 114.2 \\ \hline
    \ilmIdxEncAvg  &  0.22 & 0.20 & 13.1 & 12.3 & 15.0 & \phantom{1}65.1 \\ \hline
    \ilmIdxCtxAvg  &  0.26 & 0.22 & \textbf{13.0} & 12.4 & 14.9 & \phantom{1}39.1  \\ \hline
    \ilmIdxSeqEncAvg  &  0.15 & 0.18 & 13.2 & 12.6 & 15.2 & \phantom{1}28.4 \\ \hline
    \ilmIdxTrainedLstm & 0.22 & 0.20 & \textbf{13.0} & \textbf{12.2} & \textbf{14.8} & \phantom{1}25.6 \\ \hline
    \end{tabular}
    }
\end{table}

\begin{table}[h]
    \caption{WERs [\%] on LibriSpeech 960h of AED model with Shallow Fusion (SF),
    Density Ratio (DR), and the proposed ILM estimation methods.
    ILM BPE-level perplexity (PPL) is reported on dev-other.}
    \label{tab:libri-lstm-ilme}
    \centering
    \resizebox{\columnwidth}{!}{
    \begin{tabular}{|l|c|c|c|c|c|c|c|}
    \hline
    Method  & \shortstack{$\lambda_1$ \\ (LM)} & \shortstack{$\lambda_2$ \\ (ILM)} & \shortstack{dev-\\clean} & \shortstack{dev-\\other} & \shortstack{test-\\clean} & \shortstack{test-\\other} & \shortstack{ILM\\PPL} \\ \hline
    None & 0.0\phantom{1}  & 0.0\phantom{1} & 3.77 & 10.37 & 4.10 & 10.88 & - \\ \hline \hline
    SF   &  0.35 & 0.0\phantom{1} & 2.68 & \phantom{1}6.80 & 2.90 & \phantom{1}7.59 & - \\ \hline
    DR & 0.40  & 0.20 & 2.40 & \phantom{1}6.68 & \textbf{2.56} & \phantom{1}7.22 & - \\ \hline \hline
    train w. LM & 0.54 & 0.0\phantom{1} & 2.36 & \phantom{1}6.19 & 2.58 & \phantom{1}6.81 & - \\ \hline\hline
    \ilmIdxZero &  0.52 & 0.28 & 2.61 & \phantom{1}6.43 & 2.88 & \phantom{1}6.96 & 741 \\ \hline
    \ilmIdxEncAvg & 0.50 & \multirow{3}{*}{0.30} & 2.35 & \phantom{1}6.19 & 2.71 & \phantom{1}6.76 & 543  \\ \cline{1-2} \cline{4-8}
    \ilmIdxCtxAvg & 0.46 & & \textbf{2.33} & \phantom{1}6.19 & 2.63 & \phantom{1}6.74 & 662 \\ \cline{1-2} \cline{4-8}
    \ilmIdxSeqEncAvg  & 0.48 & & 2.44 & \phantom{1}6.34 & 2.78 & \phantom{1}7.01 & 274  \\ \hline
    \ilmIdxTrainedLstm & 0.64 & 0.42 & 2.37 & \phantom{1}\textbf{5.76} & 2.64 & \phantom{1}\textbf{6.53} & 311 \\ \hline
    \end{tabular}
    }
\end{table}

\begin{table}[h]
  \caption{WERs [\%] on LibriSpeech 960h of AED model evaluated on out-of-domain
  TED-LIUM-V2 dev and test sets.
  Results are with Shallow Fusion (SF), Density Ratio (DR), and the proposed
  ILM estimation methods.}
  \label{tab:cross-lstm}
  \centering
  \begin{tabular}{|l|c|c|c|c|}
  \hline
  Method & \shortstack{$\lambda_1$ \\ (LM)} & \shortstack{$\lambda_2$ \\ (ILM)} & TLv2-dev & TLv2-test \\ \hline
  None & 0.0\phantom{1}  & 0.0\phantom{1} & 22.0 & 22.9 \\ \hline \hline
  SF   & 0.50  & 0.0\phantom{1} & 18.5 & 19.3  \\ \hline
  DR & 0.62 & 0.48 & 16.6 & 17.8 \\ \hline \hline
  \ilmIdxZero & 0.54  & 0.36 & 17.3 & 18.3 \\ \hline
  \ilmIdxEncAvg  & 0.56 & 0.68 & 16.7 & 17.5 \\ \hline
  \ilmIdxCtxAvg  & 0.46 & 0.58 & 16.8 & 18.0 \\ \hline
  \ilmIdxSeqEncAvg & 0.52 & 0.60 & 16.7 & 18.0 \\ \hline
  \ilmIdxTrainedLstm & 0.68 & 0.58 & \textbf{16.1} & \textbf{16.9} \\ \hline
\end{tabular}
\end{table}

\subsection{Training with External LM}
We include our result for the AED model trained together with an
external LM in \Cref{tab:libri-lstm-ilme} with the ILM estimation methods.
We observe that we achieve similar WER as the average-based ILM estimation
methods without further prior correction.
Also, the optimal LM-scale is higher compared to Shallow Fusion and more in line
with the correction-based methods.

\subsection{Under-parameterized Decoder}

In \Cref{tab:swb-small-dec}, we show results for models trained with
under-parameterized decoders by reducing the LSTM dimension on both
Switchboard and LibriSpeech datasets.
We can observe that the relative improvement after adding an external LM
is increased when the decoder contains fewer parameters.
However,
the final WER is still worse than the baseline with 1000 LSTM units.

\begin{table}[h]
  \caption{
  WERs [\%] of different decoder LSTM dimensions on Switchboard 300h
  and LibriSpeech 960h.
  Shallow Fusion is used.}
  \label{tab:swb-small-dec}
  \centering
  \begin{tabular}{|c|c||c||c|c|}
    \hline
    LSTM &   & SWB & \multicolumn{2}{c|}{{LibriSpeech}} \\
     Dim. & LM & Hub5'00 & dev-clean & dev-other \\ \hline

    \multirow{2}{*}{1000} & No & 13.8 & 3.77 & 10.37 \\ \cline{2-5}
                          & Yes & \textbf{13.5} & 2.68 & \phantom{1}\textbf{6.81} \\ \hline \hline


    \multirow{2}{*}{500} & No & 14.2 & 4.10 & 11.28 \\  \cline{2-5}
                        & Yes & 13.7 & \textbf{2.65} & \phantom{1}7.48 \\ \hline \hline

    \multirow{2}{*}{300} & No & 14.3 & 4.06 & 11.07\\  \cline{2-5}
                         & Yes & 13.8 & 2.69 & \phantom{1}7.29 \\ \hline \hline


    \multirow{2}{*}{100} & No & 14.8 & 4.28 & 11.66\\ \cline{2-5}
                         & Yes & 14.3 & 2.73 & \phantom{1}7.57 \\ \hline 

\end{tabular}
\end{table}

%
%
%
%
%

\subsection{Limited Context Decoder}

\begin{table}[h]
    \caption{WERs [\%] on Switchboard 300h using FF decoder AED model with
    Shallow Fusion (SF), Density Ratio (DR), and the proposed ILM estimation
    methods.
    ILM BPE-level perplexity (PPL) is reported on Hub5'00.}
    \label{tab:swb-ff-dec}
    \centering
    \resizebox{\columnwidth}{!}{
    \begin{tabular}{|l|c|c|c|c|c|c|}
    \hline
    Method & \shortstack{$\lambda_1$\\(LM)}  & \shortstack{$\lambda_2$\\(ILM)}  & Hub5'00 & Hub5'01 & RT03  & \shortstack{ILM\\PPL}  \\ \hline
    None               & 0.0\phantom{1}   & 0.0\phantom{1}               & 14.6         & 14.0 & 16.8 & -       \\ \hline \hline
    SF                 & 0.22             & 0.0\phantom{1}               & 13.8         & 13.2 & 15.6 & -       \\ \hline
    DR                 & 0.21             & 0.04                         & 13.7         & 13.2 & 15.6 & - \\ \hline \hline
    \ilmIdxZero        & \multirow{2}{*}{0.29}  & 0.25                   & 13.2         & 12.6 & 15.0 & 68  \\ \cline{1-1} \cline{3-7}
    \ilmIdxEncAvg      &                  & 0.23                & \textbf{13.0} & \multirow{2}{*}{\textbf{12.4}} & \textbf{14.8} & 47   \\ \cline{1-4} \cline{6-7}
    \ilmIdxCtxAvg      & \multirow{2}{*}{0.27} & \multirow{2}{*}{0.25}  & 13.2         &  & \multirow{3}{*}{14.9} & 58  \\ \cline{1-1} \cline{4-5} \cline{7-7}
    \ilmIdxSeqEncAvg   &                  &                              & 13.3         & 12.5 & & \multirow{2}{*}{28}  \\ \cline{1-3} \cline{4-5}
    \ilmIdxTrainedLstm & 0.29             & 0.29                         & 13.1         & 12.6 &  &       \\ \cline{1-1} \cline{3-7} \hline
    \end{tabular}
    }
\end{table}

\begin{table}[h]
    \caption{WERs [\%] on LibriSpeech 960h using FF decoder AED model with
    Shallow Fusion (SF), Density Ratio (DR), and the proposed ILM estimation
    methods.
    ILM BPE-level perplexity (PPL) is reported on dev-other.}
    \label{tab:libri-ff-dec}
    \centering
    \resizebox{\columnwidth}{!}{
    \begin{tabular}{|l|c|c|c|c|c|c|c|}
    \hline
    Method & \shortstack{$\lambda_1$\\(LM)} & \shortstack{$\lambda_2$\\(ILM)} & \shortstack{dev-\\clean}
    & \shortstack{dev-\\other}  & \shortstack{test-\\clean} & \shortstack{test-\\other} & \shortstack{ILM\\PPL}  \\ \hline
    None    & 0.0\phantom{1}       & 0.0\phantom{1}       & 4.21          & 11.66                     & 4.30 & 12.25& -      \\ \hline \hline
    SF                 & 0.35      & 0.0\phantom{1}       & 2.77          & \phantom{1}7.94           & 3.05 & \phantom{1}8.04 & -      \\ \hline
    DR                 & 0.58      & 0.30                 & 3.14          & \phantom{1}7.18           & 3.30 & \phantom{1}8.31 & -      \\ \hline \hline
    \ilmIdxZero        & 0.43      & 0.28                 & 2.81          & \phantom{1}7.31           & 3.01 & \phantom{1}8.06 & 795    \\ \hline
    \ilmIdxEncAvg      & 0.44      & \multirow{2}{*}{0.42}& \textbf{2.54} & \phantom{1}6.73   & \textbf{2.81} & \phantom{1}7.21 & 462    \\ \cline{1-2} \cline{4-8}
    \ilmIdxCtxAvg      & 0.50      &                      & 2.96          & \phantom{1}7.16           & 3.25 & \phantom{1}7.93 & 921    \\ \hline
    \ilmIdxSeqEncAvg   & 0.48      & 0.46                 & 2.74          & \phantom{1}6.60           & 2.87 & \phantom{1}\textbf{7.10} & 267    \\ \hline
    \ilmIdxTrainedLstm & 0.64      & 0.52                 & 2.80          & \phantom{1}\textbf{6.58}  & 3.07 & \phantom{1}7.36 & 196 \\ \hline
    \end{tabular}
    }
\end{table}


We replace the recurrent LSTM decoder from \Cref{att:query} with a FF decoder
in \Cref{eqn:FFDecoder}. The motivations behind this are:
(1) AED model might not need the full label context,
(2) limiting the label context can reduce the decoder LM effectiveness.
We use 1 linear projection layer with bias followed by tanh activation function.
We notice that, under this configuration, a context size of 3 is optimal
($\mathrm{k}=3$ in \Cref{eqn:FFDecoder}).
Experiments on Switchboard and LibriSpeech datasets are
shown in \Cref{tab:swb-ff-dec,tab:libri-ff-dec}.
We observe that LM integration is needed otherwise the model is worse
and that the relative improvement is stronger
compared to the LSTM decoder. 
On Switchboard, after we applied our ILM estimation methods,
the results became quite comparable to using the LSTM decoder.
On LibriSpeech, with LSTM decoder the performance is still better but we
argue that there is a possibility for more improvement with careful training
and tuning.

\section{Conclusion}

In this work, we evaluated different methods to estimate the internal
language model (ILM) from the decoder of attention-based encoder-decoder (AED)
models.
We proposed several new methods that outperformed all of the existing approaches.
We presented a simple method for ILM estimation, called \ilmIdxTrainedLstm,
that achieved the best results on both intra-domain and cross-domain evaluations.
We also showed that, in general, it is better to use encoder bias information
instead of masking out the input representation for the ILM estimation.
We showed that it is possible to train a feed-forward or limited context decoder
for an AED model achieving comparable results to a recurrent decoder on Switchboard.
This can have further applications, for example, for lattice recombination.
Moreover, we showed that the capacity of the decoder can be reduced without a huge
degradation in word error rate.

\section{Acknowledgments}
This project has received funding from the European Research Council (ERC)
under the European Union’s Horizon 2020 research and innovation programme
(grant agreement n\textsuperscript{o}~694537, project "SEQCLAS").
The work reflects only the authors' views and the European Research
Council Executive Agency (ERCEA) is not responsible for any use
that may be made of the information it contains.
This work was partly funded by the Google Focused Award
"Pushing the Frontiers of ASR: Training Criteria and Semi-Supervised Learning".
This work was partially supported by the project HYKIST funded by
the German Federal Ministry of Health on the basis of a decision of the
German Federal Parliament (Bundestag).

\bibliographystyle{IEEEtran}

\bibliography{paper}

\end{document}